\documentclass[runningheads]{llncs}

\usepackage{eccv}

\usepackage{eccvabbrv}

\usepackage{graphicx}
\usepackage{booktabs}

\usepackage[accsupp]{axessibility}  
\usepackage{subcaption}
\usepackage[table]{xcolor}
\usepackage{pifont}
\usepackage{amsmath}
\usepackage{xcolor}
\usepackage{amssymb}
\usepackage{amsfonts}
\usepackage{tcolorbox}
\usepackage{multirow}

\usepackage{hyperref}

\usepackage{orcidlink}
\usepackage{marvosym}

\begin{document}

\title{Unleashing Multimodal Large Language Models for Training-free HOI Detection in the Wild} 

\titlerunning{Unleashing MLLMs for Training-free HOI Detection in the Wild}




 
\author{Ting Lei\textsuperscript{*} \and
Jialin Liu\textsuperscript{*} \and
Zhu Xu\and
Yuxin Peng \and
Yang Liu \textsuperscript{\Letter}}

\authorrunning{T.~Lei et al.}

\institute{Wangxuan Institute of Computer Technology, Peking University, Beijing, China\\
\email{ting\_lei@pku.edu.cn, ljl@stu.pku.edu.cn, xuzhu@stu.pku.edu.cn, pengyuxin@pku.edu.cn, yang\_liu@pku.edu.cn}}

\maketitle

\begingroup
\renewcommand{\thefootnote}{}
\footnotetext{\textsuperscript{*} Equal contribution. \quad
\textsuperscript{\Letter} Corresponding author.}
\endgroup

\begin{abstract}

Human-object interaction detection (HOID) has traditionally been formulated as a supervised detection problem over predefined interaction categories. While such paradigms achieve strong performance on closed-set benchmarks, they fundamentally entangle interaction understanding with dataset-specific supervision, limiting their ability to generalize to open-world and compositional scenarios. 
Recent HOI detectors attempt to leverage MLLMs through prompting strategies to transfer interaction-specific knowledge. However, such prompt-based approaches primarily focus on extracting discriminative representations from pretrained models, while underexploring their inherent multimodal reasoning capabilities. As a result, they struggle to provide informative contextual reasoning for ambiguous and open-world interaction scenarios.
In this work, we present AgentHOI, a training-free, agentic framework that transfers the generalist multimodal reasoning capabilities of foundation models to HOI detection in the wild. Instead of learning interaction classifiers, AgentHOI modularly orchestrates complementary vision foundation modules to perform open-ended semantic reasoning and spatial grounding in a coordinated manner.
To address the challenges of incomplete interaction discovery and ambiguous localization in complex scenes, we introduce two key mechanisms: (1) Context-aware Multi-round Reasoning, which progressively refines interaction hypotheses to ensure exhaustive and compositional HOI discovery, and (2) Multifaceted Interaction Localization, which enhances grounding precision by generating instance-specific descriptions that integrate semantic, spatial, and appearance cues.
Extensive experiments demonstrate that AgentHOI achieves superior performance over state-of-the-art supervised and weakly supervised methods in real-world settings, despite requiring no HOID data for training. Code is available at \url{https://github.com/oceanflowlab/AgentHOI}.

\end{abstract}

\section{Introduction}

Human-object interaction detection (HOID) aims to identify and localize structured triplets in the form of $\langle$human, action, object$\rangle$ from visual scenes~\cite{chao2018HICO-DET,wang2021SWIG-HOI}. Existing approaches predominantly formulate HOID as a supervised detection problem over predefined interaction vocabularies. Under this paradigm, models are trained to learn interaction-specific classifiers from large-scale annotated datasets and achieve impressive results on closed-set benchmarks~\cite{zheng2026omnivtg,yin2025toolvqa,zhang2023pvic,wu2024pose_aware,hui2025image,xu2025trkt}. However, such methods tightly couple interaction understanding with dataset-specific supervision, making them inherently limited in open-vocabulary and open-world scenarios where interaction categories and visual distributions cannot be exhaustively enumerated. In this paper, we use \emph{open-vocabulary} to denote generalization to unseen HOI classes, while \emph{open-world} further includes robustness to unseen image styles and visual distributions.

\begin{figure}[t!]
    \centering
    \includegraphics[width=0.7\textwidth]{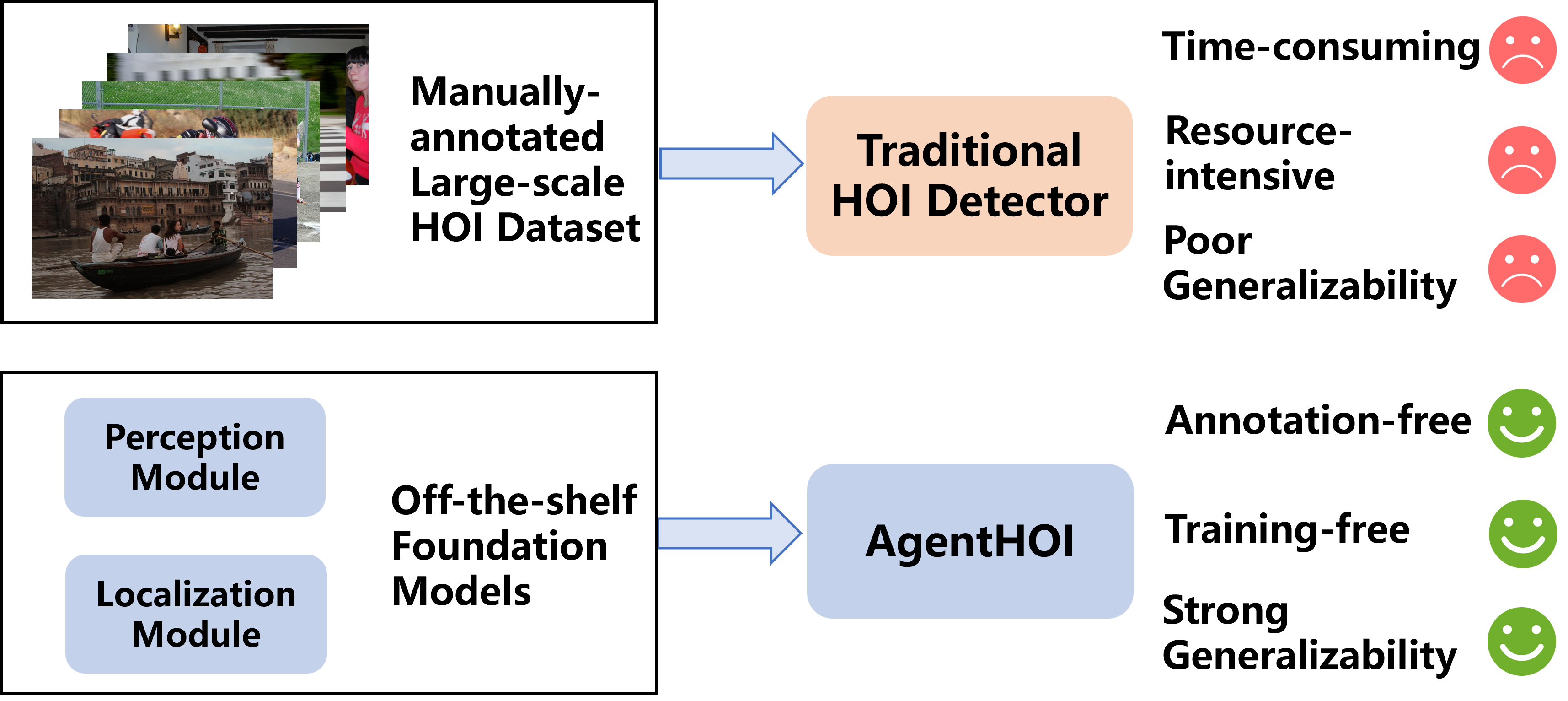}
    \caption{Framework comparison with previous methods. Most existing methods depend on manually annotated, large-scale HOI datasets for training, which is both time-consuming and resource-intensive. Moreover, they often exhibit limited generalizability in real-world settings. In contrast, our AgentHOI framework requires no annotation or training, offering strong generalizability by leveraging off-the-shelf vision foundation models.}
    \label{fig:teaser}
    \vspace{-2em}
\end{figure}

To relax the restriction of closed interaction vocabularies, recent studies have explored vision-language foundation models and MLLMs~\cite{peng2026survey,yang2026gala,achiam2023gpt,bai2025qwen2-5,chen2025helmet} for HOI detection~\cite{yang2024MP_HOI,guo2024HOIGen,lei2025hola,gao2026taming,lei2025INP-CC}. A common strategy is to introduce textual prompts, ranging from hand-crafted templates such as ``a photo of a person [action] a/an [object]''~\cite{liao2022gen,ADA_CM,qu2022distillation,iftekhar2022look,xu2024semantic} to learnable or LLM-generated descriptions~\cite{ting2024CMMP,lei2025hola,lei2024CMD-SE,cao2023UniHOI,luo2024SIC,liu2026confidence,gao2025identity}. These prompts align interaction categories with pretrained vision-language representations and improve recognition of unseen HOI classes. Nevertheless, most existing methods still follow a supervised detection paradigm, where prompts are used to enhance task-specific classifiers trained on annotated HOI datasets rather than to perform open-ended interaction reasoning. As a result, their behavior remains coupled to predefined label spaces and dataset-specific visual distributions, limiting robustness in open-world scenarios where interaction compositions, object appearances, and image styles may differ substantially from the training data. Moreover, although MLLMs possess strong multimodal reasoning capabilities, they are mainly used as feature extractors, semantic aligners, or text generators, leaving their potential for discovering and disambiguating complex interactions underexplored.

To address the aforementioned challenges, we introduce AgentHOI, a training-free, agentic framework for HOI detection in the wild. Rather than learning interaction classifiers from annotated HOI datasets, AgentHOI orchestrates complementary vision foundation modules to perform two coordinated processes: open-ended semantic reasoning and precise spatial grounding. The perception module, built upon MLLMs, infers plausible interaction triplets through compositional reasoning, while the localization module grounds descriptive phrases to spatial coordinates. By coordinating these modules in an agent-like manner, AgentHOI transfers the generalist reasoning capabilities of foundation models to structured HOI detection.

\begin{figure}[t!]
   \centering
    \begin{subfigure}{0.41\textwidth}
        \includegraphics[width=\textwidth]{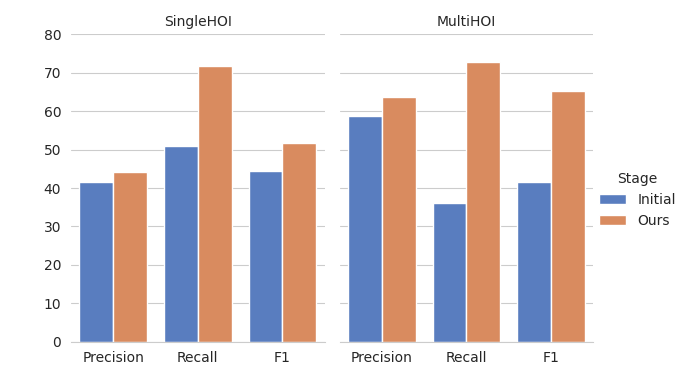}
        \caption{Performance comparison of context-aware multi-round reasoning under SingleHOI and MultiHOI scenes.}
        \label{fig:metric2round}
    \end{subfigure}
    \hfill
    \begin{subfigure}{0.55\textwidth}
        \begin{minipage}{\textwidth}
            \includegraphics[width=0.48\textwidth]{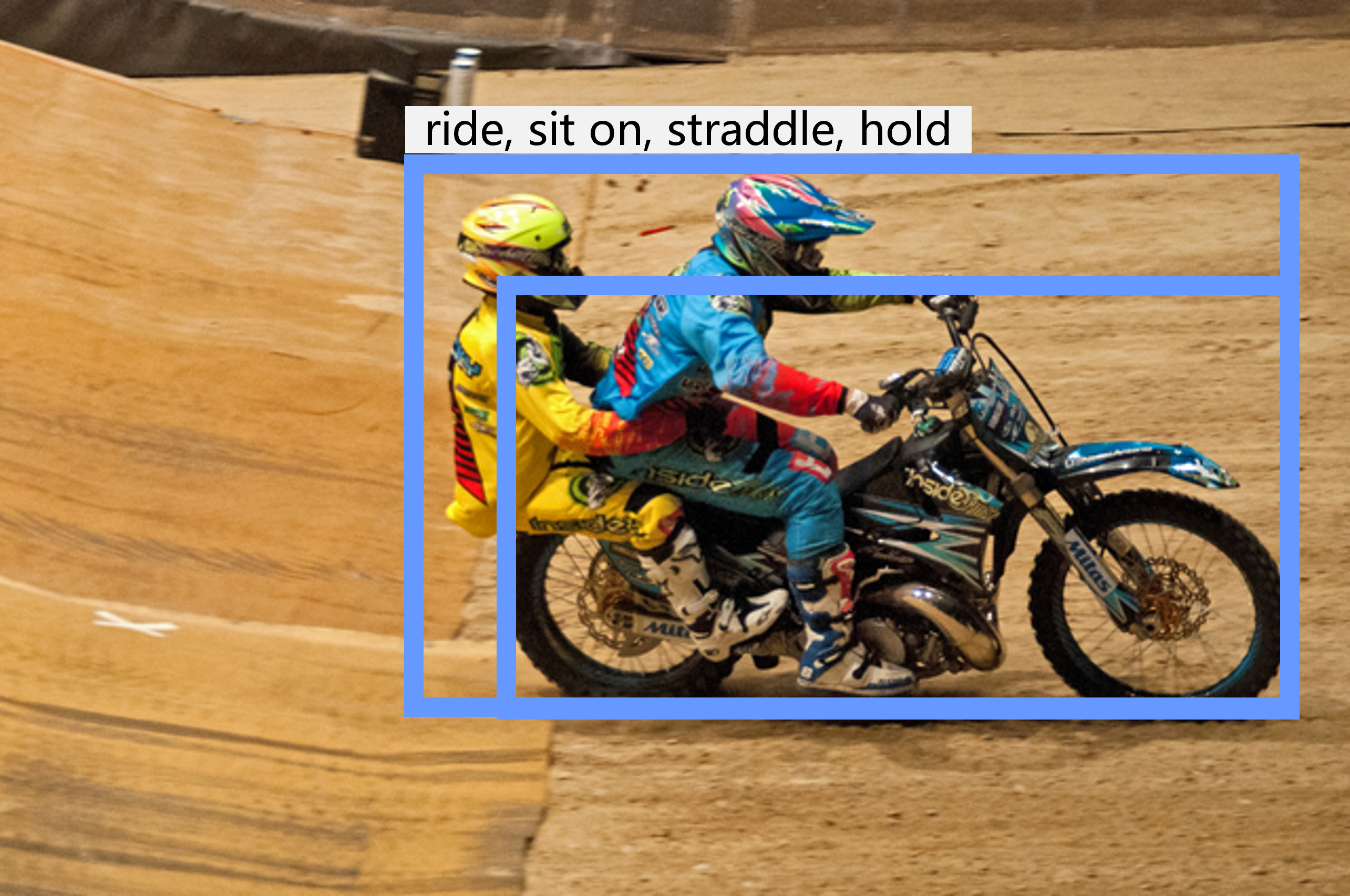}
            \includegraphics[width=0.48\textwidth]{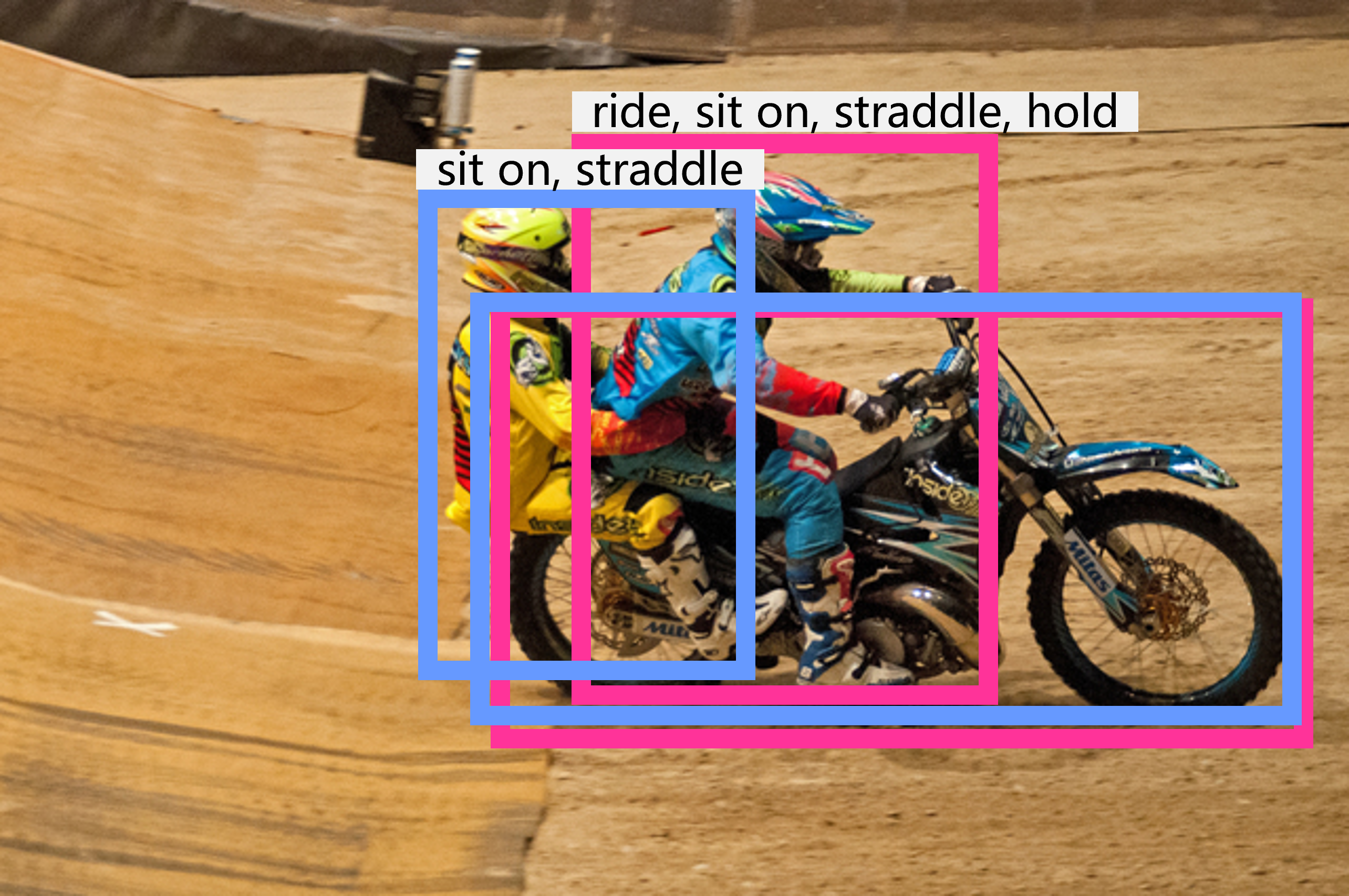}
        \end{minipage}
        \caption{Interaction localization improvement via multifaceted phases.}
        \label{fig:query_images}
    \end{subfigure}
    \caption{(a) The context-aware, multi-round reasoning mechanism enhances interaction recognition, particularly for images with multiple HOIs~(right). Initial: Single-round answer by MLLM. (b) Left: without multifaceted interaction localization (MIL). Right: with MIL. The multifaceted interaction localization enhances the accuracy of localizing human-object pairs in scenes with multiple similar HOI concepts.}
    \label{fig:comparison}
    \vspace{-1em}
\end{figure}

Despite this collaborative setup, two core challenges remain:
(1) Incomplete interaction discovery: The perception module may miss interactions in cluttered or multi-action scenes, often focusing on only the most salient triplets. For instance, a person may interact with multiple objects simultaneously, or a single human-object pair may involve multiple actions. This complexity frequently leads MLLMs to miss valid HOIs. As shown in Fig.~\ref{fig:metric2round}, interaction recall and F1 score drop by 14.92\% and 2.83\%, respectively, in scenes with multiple HOI triplets compared to single-HOI scenes.
(2) Ambiguous grounding: In scenes with visually similar instances or repeated interactions, the localization module struggles to disambiguate when prompted with generic or underspecified phrases. As illustrated in Fig.~\ref{fig:query_images}, grounding errors occur when multiple individuals interact similarly with the same object (e.g., different people sitting on the same motorbike).

To address these challenges, we introduce two general mechanisms that enhance the robustness and completeness of foundation-model-driven HOI detection.
(1) \textbf{Context-aware Multi-round Reasoning} progressively refines interaction hypotheses through staged reasoning. Specifically, the perception module iteratively identifies salient interactions, mines additional candidates conditioned on contextual anchors, and reassigns potentially missing verbs based on fine-grained action priors. This process improves interaction completeness in compositional scenes.
(2) \textbf{Multifaceted Interaction Localization} enhances grounding precision by generating instance-specific descriptions that combine semantic labels with spatial and appearance cues. By enriching textual representations before spatial grounding, the framework enables more reliable disambiguation among visually similar entities.
Together, these mechanisms transform HOI detection from a fixed-vocabulary detection and classification task into a modular reasoning-and-grounding process. AgentHOI requires no HOI-specific training data and no parameter updates, yet achieves strong compositional generalization in open-world settings.

Our contributions are summarized as follows:

\begin{itemize}
    \item We propose AgentHOI, the first agentic framework that reformulates HOI detection as a structured reasoning-and-grounding problem, enabling training-free HOI detection in the wild.
    \item We introduce Context-aware Multi-round Reasoning and Multifaceted Interaction Localization, two plug-and-play mechanisms that improve interaction completeness and grounding disambiguation under complex real-world conditions.
    \item Extensive experiments demonstrate that AgentHOI achieves superior performance over state-of-the-art supervised and weakly supervised methods in real-world settings, despite requiring no HOID data for training.
\end{itemize}

\section{Related Work}
\label{sec:related}

\noindent\textbf{Generic HOI Detection.}
HOI Detection aims to detect all the HOI instances within images, including locating the interactive human-object pair and recognizing their interaction categories. The majority approaches~\cite{cao2023RmLR,zhang2022UPT,gao2020drg,wang2024bilateral,jiang2024SCTC,luo2024SIC,wang2022distance,Kim_2023_CVPR,li2023sqab,cheng2024parallel,geng2025horp} for HOID are fully-supervised, which requires the human-object pair bounding boxes and interaction categories as annotation. Though a great progress, there exists a significant reliance 
on large-scale high-quality HOI annotations to enable accurate HOI detection, which is time-consuming high-cost to collect. Besides, complex scenes, i.e., images with multiple people and objects, further aggravated the annotation burden. To alleviate it, weakly-supervised HOID approaches~\cite{wan2023weaklyHOI,unal2023weaklysupervised} are proposed, which only require image-level interaction category annotations without bounding boxes for model training. However, they still require manual HOI annotations and remain constrained by predefined vocabularies and dataset-specific distributions, limiting their generalization to unseen interaction compositions in real-world settings.
In contrast, we revisit HOI detection from a foundation-model perspective. Instead of learning task-specific interaction classifiers, our method leverages off-the-shelf vision foundation models and reformulates HOI detection as structured reasoning and grounding. The resulting framework is entirely training-free and aims to enable more scalable and real-world HOI understanding.

\noindent\textbf{Open Vocabulary HOI Detection.}
Open vocabulary HOI Detection (OV-HOID) aims to detect interactions unseen in the training set, which plays crucial roles in developing practical HOI detection models working in real-world scenarios. 
Early works such as VCL~\cite{hou2020VCL} and FCL~\cite{hou2021FCL} combine object representations and human representations to compose unseen HOI samples for learning.
ATL~\cite{hou2021ATL} exploits extra object
datasets to discover knowledge for unseen HOI categories. 
More recently, advancements in CLIP-wise discriminative vision-language models (VLM) show outstanding generalization and promising zero-shot vision-text alignment capability, which facilitates the further development of OVHOID models~\cite{liao2022gen,ning2023hoiclip,li2024logichoi,mao2023clip4hoi,yang2024CaCLIP,wu2022EoID,lin2025sgc}.
EoID~\cite{wu2022EoID} distills action probability distribution from CLIP to enable unseen action recognition, GEN-VLKT~\cite{liao2022gen} and HOICLIP~\cite{ning2023hoiclip} utilize CLIP text embeddings to initialize the action classifier and CLIP visual features to guide the interactive representation learning. CMD-SE~\cite{lei2024CMD-SE} leverages body part descriptions to help distinguish fine-grained interactions. MP-HOI~\cite{yang2024MP_HOI} optimizes the HOI task as a similarity learning process using a unified contrastive loss to learn generalizable object/interaction representations from large-scale data.
However, existing CLIP-based OV-HOID methods still rely on supervised learning, requiring substantial annotated data for supervision and extensive training to achieve competitive performance. In contrast, our AgentHOI departs from classifier-based learning and leverages off-the-shelf vision foundation models for interaction reasoning. By reformulating HOI detection as a structured reasoning and grounding process, our framework eliminates task-specific training altogether and moves toward truly training-free, open-world HOI understanding.

\section{Method}
\label{sec:method}

In this section, we begin by outlining the problem formulation. Next, we introduce AgentHOI-Base, a training-free framework that leverages vision foundation models to address the limitations of existing HOI detection methods, particularly their dependence on large annotated datasets and lack of generalizability to real-world settings.
To better tackle complex scenes, we incorporate two key components: a context-aware multi-round reasoning mechanism, which iteratively refines detection results through progressive querying, and a Multifaceted Interaction Localization mechanism, designed to reduce ambiguity in interaction localization. Together, these modules constitute the complete AgentHOI system.

\begin{figure*}[!t]
    \centering
    \begin{subfigure}[b]{1.0\textwidth}
      \centering
      \includegraphics[width=0.9\textwidth]{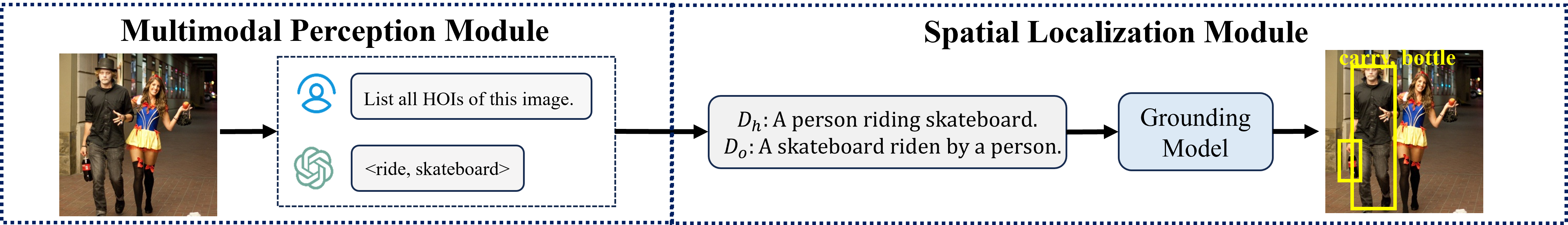}
      \caption{The overall framework of AgentHOI-Base.}
      \label{fig:baseline}
    \end{subfigure}
    \hfill
    \begin{subfigure}[b]{1.0\textwidth}
      \centering
      \includegraphics[width=0.9\textwidth]{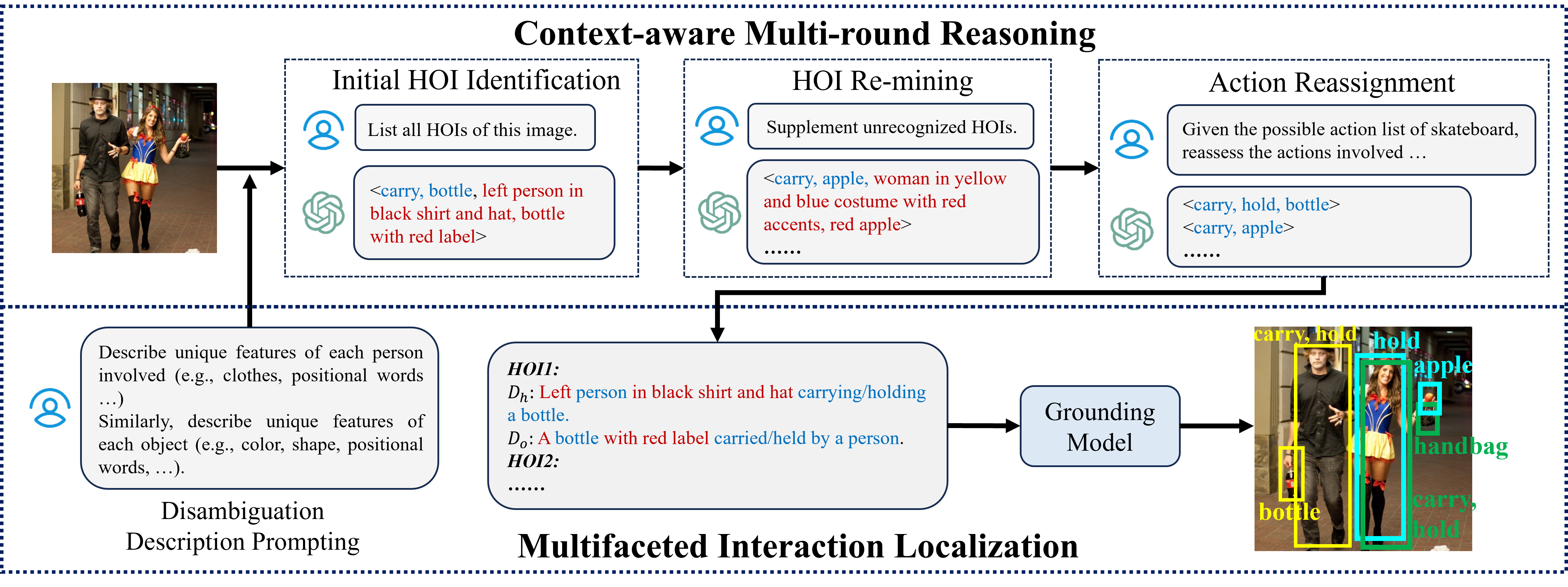}
      \caption{The overall framework of AgentHOI.}
      \label{fig:framework}
    \end{subfigure}
    \caption{(a) We first propose a training-free generalist vision agent baseline, AgentHOI-Base, consisting of a multimodal perception module and a spatial localization module for interaction category identification and target localization, respectively. (b) 
    Our AgentHOI further includes two components: Context-aware Multi-round Reasoning and Multifaceted Interaction Localization. We first identify potential interactions with the MLLM. In the HOI-re-mining phase, we re-query the MLLM using identified interactions as anchors to detect missed ones. In the action reassignment phase, we apply a fine-grained prior to refine the action categories for interactive human-object pairs. To minimize ambiguity during grounding, we introduce a disambiguation description strategy, prompting the MLLM to generate detailed descriptions of human-object pairs, which helps the grounding model distinguish similar instances. Black text denotes input prompts for MLLM, blue text denotes mined or refined HOI triplets, and red text indicates disambiguation cues for grounding.}
    \label{fig:a+b}
\end{figure*}


\subsection{Problem Formulation}
Given an image \( I \in \mathbb{R}^{H \times W \times 3} \), the task is to detect all human-object interactions (HOIs) within it. Each interaction is a tuple \((b_h, b_o, c_o, c_a)\), where \( b_h, b_o \in \mathbb{R}^4 \) are bounding boxes for the human and object, \( c_o \in \mathcal{C}_o = \{ c_o^1, \dots, c_o^N \} \) is the object category, and \( c_a \in \mathcal{C}_a = \{ c_a^1, \dots, c_a^M \} \) is the associated action. The output is a set \( \mathcal{T} = \{ (b_h^i, b_o^i, c_o^i, c_a^i) \}_{i=1}^{K} \) of \( K \) predicted HOIs in the image.
We consider the training-free setting, where the model undergoes no task-specific fine-tuning.

\subsection{AgentHOI-Base}
To overcome the limitations of traditional methods—such as reliance on large annotated datasets, extensive training times, and poor generalization—we propose AgentHOI-Base, a training-free HOI detection baseline that leverages the complementary capabilities of vision foundation models. As illustrated in Fig.~\ref{fig:baseline}, this baseline consists of two sequential modules: a \textit{Multimodal Perception Module} via an MLLM, and a \textit{Spatial Localization Module} via a grounding model.

In the first stage, we employ an MLLM \( \mathcal{M}_L \) to identify HOI categories from an image \( I \in \mathbb{R}^{H \times W \times 3} \) guided by an instruction prompt \( P_1 \), which includes prior object \( \mathcal{C}_o \) and action \( \mathcal{C}_a \) sets. The output is a set of action-object pairs \( \{(c_a, c_o)\} \), generated based on the model’s open-world knowledge without task-specific training.
Next, a grounding model \( \mathcal{M}_g \) localizes these interactions. For each pair \( (c_a, c_o) \), we form phrases like ``a person \( c_a \) an \( c_o \)'' and ``a \( c_o \) \( c_a \)-ed by a person'' to retrieve the human and object boxes \( b_h \) and \( b_o \).
Overall, AgentHOI-Base provides a training-free pipeline for in-the-wild HOI detection by leveraging the complementary strengths of generalist models.

\subsection{Context-aware Multi-round Reasoning} 
\label{subsec:multiround}

Identifying HOIs accurately within images can be challenging, particularly in complex scenes with multiple, simultaneous interactions. The above AgentHOI-Base often struggle to recognize all interactions, especially when the scene contains multiple interactive human-object pairs. To address this, we propose a context-aware multi-round reasoning mechanism leveraging the MLLM to enhance AgentHOI’s perception. As shown in the upper part of Fig.~\ref {fig:framework}, this mechanism refines interaction recognition through multiple query rounds. The process has three stages: (1) initial HOI identification, (2) HOI re-mining, and (3) action reassignment, each refining the results further.

\noindent \textbf{Initial HOI Identification.} 
The first phase of the mechanism focuses on identifying all potential interactions in the image. By utilizing the prompt \( P_1 \) introduced in the previous subsection, the MLLM generates an initial set of interaction categories, represented as action-object pairs without bounding boxes, denoted \( T_1 \):
\begin{equation}
    T_1 = \mathcal{M}_{L}(I, P_1)
\end{equation}
where \( T_1 = \{(c_a,p_a,c_o,p_o)\}_{i=1}^{n_1} \), with \(c_a\) and \(c_o\) denoting the action and object categories, and \(p_a, p_o\) their confidence scores.

\noindent \textbf{HOI Re-mining.}
Due to the complexity of HOI scenes, the initial round may miss some interactions, especially when multiple interactions occur simultaneously in the same image. For instance, when a person is interacting with several objects at once, the MLLM may prioritize the most salient interaction, potentially overlooking others. As shown in Fig.~\ref {fig:framework}, the first round only identifies \textit{ride skateboard} while neglecting \textit{carry backpack}, requiring a second chance to discover the missing HOI. To mitigate this, we introduce the HOI re-mining phase. In this phase, we query the MLLM again, using the previously identified interactions \( T_1 \) as anchors.
These anchors help guide the MLLM to reason about any additional interactions that might have been missed. The MLLM then identifies additional HOIs, producing the supplemented set \( T_2 \):
\begin{equation}
    T_2 = \mathcal{M}_L(I, P_2, T_1)
\end{equation}
Here, \( P_2 \) is an adapted prompt that incorporates \( T_1 \) as contextual prior, encouraging the model to explore complementary interactions.
Specifically, as illustrated in the upper-middle of Fig.~\ref {fig:framework}, in the second query, the MLLM newly recognizes the interaction \textit{wear backpack}, which had been overlooked initially.
The output is thus an additional set of HOIs \( T_2 = \{(c_a, p_a, c_o, p_o)\}_{i=1}^{n_2} \), enriching the coverage of interaction categories. This iterative querying strategy allows the MLLM to move beyond its first-pass biases and uncover overlooked triplets.

\noindent \textbf{Action Reassignment.}
In some instances, the same human-object pair may involve multiple potential actions. To handle this, we introduce a fine-grained action prior associated with each object to guide the MLLM in reassessing the action categories of previously identified interactions~\cite{liu2024freea}. Formally, for each object category \( c_o \), we provide a set of possible actions \( d_o \) and incorporate this prior into the reassignment prompt \( P_3 \). This enables the MLLM to refine the action categories of previously identified HOIs:
\begin{equation}
    T_{3} = \mathcal{M}_{L}(I, P_3, \{T_1, T_2\})
\end{equation}
where \( T_{3} = \{(c_a^1, \dots, c_a^{k_o}), (p_a^1, \dots, p_a^{k_o}), c_o, p_o\}_{i=1}^{n_1 + n_2} \), with \( k_{c_o} \) denoting the number of plausible actions for object \( c_o \). 
As illustrated in Fig.~\ref{fig:framework}, this step allows the model to further refine interaction semantics, generating instances such as stand on skateboard, which was missed or unclear in earlier rounds.

\subsection{Multifaceted Interaction Localization}
\label{subsec:multifaceted}

While the enhanced perception module of AgentHOI exhibits strong reasoning capabilities in identifying interactions, it often struggles with precise spatial localization. To address this limitation, we propose a \textit{Multifaceted Interaction Localization} mechanism that complements the semantic reasoning of MLLMs with the spatial grounding ability of specialized grounding models. By leveraging richly descriptive language cues, this mechanism enables accurate localization of interacting human-object pairs.

\noindent \textbf{Disambiguation via Descriptive Prompting.}  
Directly converting interaction triplets into simple templates (e.g., ``a person $<$action$>$ a $<$object$>$'') often results in ambiguity, especially in scenes with multiple similar objects or individuals. For instance, as illustrated in Fig.~\ref{fig:query_images}, if two people are riding motorcycles, the grounding model may struggle to differentiate between them.

To address this, we propose \textit{Descriptive Prompting} to enhance disambiguation. In each reasoning round \( \{P_i\}_{i=1}^3 \), prompts are extended to ask the MLLM not only for interaction categories, but also for referential descriptions of the involved human and object instances. These may include spatial, visual, or contextual cues to help differentiate similar entities. As shown in the lower-middle part of Fig.~\ref{fig:framework}, 
\( D_h \) and \( D_o \) denote descriptive features from the human and object, respectively—e.g., \( D_h \): ``Left person in black shirt and hat carrying/holding a bottle'', and \( D_o \): ``A bottle with red label carried/held by a person''.
Each interaction tuple \( \{T_{i}\}_{i=1}^{3} \) is thus extended to include human and object descriptions (\( D_h \), \( D_o \)), leading to the final result as: 
\begin{equation}
    T_{\text{final}} = \{(c_a^1, \dots, c_a^{k_o}), (p_a^1, \dots, p_a^{k_o}), c_o, p_o, D_h, D_o \}
\end{equation}

\noindent where \( (c_a^1, \dots, c_a^{k_o}) \) are the action categories associated with object \( c_o \), and \( D_h \), \( D_o \) provide descriptive attributes for the human and object respectively.

\noindent \textbf{Grounded Human-Object Localization.}  
Given the enriched interaction information, we transform each instance into two referential grounding queries for the human and object, respectively. The human description phrase \( D_h \) is constructed as ``A person that \( D_h, c_a^1, \dots, c_a^{k_o} \, c_o \)'', and the object description \( D_o \) is formulated as ``A \( c_o \) that \( D_o, c_a^1, \dots, c_a^{k_o} \) by a person''.
These descriptions are then passed into a grounding model \( \mathcal{M}_g \)~\cite{liu2023groundingdino}, which returns the bounding boxes and confidence scores:

\begin{equation}   
(b_h, s_h),\ (b_o, s_o) = \mathcal{M}_{g}(D_h),\ \mathcal{M}_{g}(D_o)
\end{equation}

\noindent where \( b_h \) and \( b_o \) denote the bounding boxes for the human and object, and \( s_h \), \( s_o \) are their respective confidence scores.
The final structured HOI output is:

\begin{equation}
\{(c_a^1, \dots, c_a^{k_o}), c_o, b_h, s_h, b_o, s_o\}
\end{equation}

\noindent To compute a confidence score \( s_{a,o}^{\text{final}} \) for final evaluation, we combine the MLLM classification confidence from the last stage of multi-round reasoning with the grounding model’s localization scores:

\begin{equation}
    s_{a,o}^{\text{final}} = s_{a,o} \cdot (s_h \cdot s_o)^{\lambda}
\end{equation}

\noindent where \( s_{a,o} = \frac{p_a + p_o}{2} \) is the averaged classification confidence, and \( \lambda \) is a balancing coefficient that modulates the influence of grounding reliability.
By unifying the high-level reasoning capabilities of MLLMs with the precise spatial predictions of grounding models through descriptive mediation, this module enables robust and accurate localization of HOIs in open-world scenarios, without reliance on annotated training data.

\section{Experiments}
\subsection{Experimental Settings}

\noindent \textbf{Dataset.}
Our experiments are conducted on the HICO-DET~\cite{chao2018learning} dataset. The HICO-DET test set contains 9,658 images and 600 HOI categories, covering 118 actions and 80 object classes. 
Following prior works~\cite{lei2025hola,ting2024CMMP}, we assess performance under four zero-shot HOI detection settings: unseen verb (UV), non-rare first unseen composition (NF-UC), rare first unseen composition (RF-UC), and unseen object (UO).

\noindent \textbf{Evaluation under Distribution Shifts.}
To further assess the robustness and real-world generalization capability of zero-shot HOI detection models, we introduce additional distribution shifts at the image level. Specifically, we apply style transfer~\cite{wu2025uso} and image quality degradation~\cite{tang2025robustr1} to the test images, simulating real-world scenarios with varying visual styles and imaging conditions. The style transfer aims to mimic domain gaps caused by artistic rendering, camera pipelines, or web image stylization, while the degradation operations simulate practical challenges such as blur, noise, and compression artifacts. These modifications create out-of-distribution test settings without altering the semantic HOI annotations, allowing us to evaluate the model’s robustness to appearance variations.
For robustness evaluation, all methods use the same transformed test images with unchanged HOI evaluation pipelines for fair comparison.

\noindent \textbf{Evaluation Metric.}
Following~\cite{ADA_CM,wan2023weaklyHOI}, we evaluate the performance using mean Average Precision (mAP). An HOI triplet prediction is considered a true positive if the Intersection over Union (IoU) between the predicted human and object bounding boxes and the corresponding ground truth boxes exceeds 0.5, and both the predicted object category and interaction label match the ground truth.

\subsection{State-of-the-art Comparison}
Tab.~\ref{tab:comparison} reports the comparison with existing zero-shot HOI detection methods on HICO-DET under three evaluation settings. To the best of our knowledge, AgentHOI is the first training-free HOI detection framework built upon a large-scale MLLM, without any task-specific fine-tuning on HOI data.
On the original test set, AgentHOI achieves the best performance on UV (29.85) and RF-UC (38.64), demonstrating strong generalization ability. The strong performance of AgentHOI on UV and RF-UC validates our reasoning-and-grounding design: context-aware multi-round reasoning helps recover interactions missed by single-pass queries, while descriptive grounding enriches triplets with instance-specific semantic, appearance, and spatial cues, enabling more reliable interaction detection.

Under distribution shifts induced by style transfer and image degradation, the superiority of AgentHOI becomes more evident. While supervised methods such as CMMP, EZ-HOI, and HOLa experience noticeable performance drops when evaluated on style-transferred and degraded images, AgentHOI maintains relatively stable performance across all unseen splits. These results highlight a key trade-off between training-free and supervised paradigms. Supervised zero-shot methods benefit from task-specific optimization and annotated human-object pairs, which can improve discrimination within the source distribution. In contrast, AgentHOI sacrifices such dataset-specific adaptation, but gains the flexibility to reason over open-ended interaction compositions without retraining or introducing HOI-specific parameters.

Under the NF-UC setting on the original test set, supervised zero-shot methods achieve slightly higher performance than AgentHOI. This can be partly explained by the characteristics of the NF-UC split: although the specific HOI triplets are unseen during training, the associated objects and actions are typically frequent in the training data, allowing supervised methods to learn strong classifiers for these components. 
Further analysis shows that many difficult NF-Unseen classes are related to \textit{no\_interaction}. Among the Bottom-20 NF-Unseen categories predicted by AgentHOI, 17 involve \textit{no\_interaction}. Prior work~\cite{zhu2025diagnosing} has shown that such categories are particularly prone to missing annotations, which can negatively affect evaluation metrics. Since AgentHOI performs open-ended reasoning without adapting to dataset-specific annotation patterns, it may predict plausible interactions that are not exhaustively labeled, leading to lower measured performance under this setting.

\begin{table}[tbp]
  \centering
  \caption{Performance comparison with state-of-the-art models of zero-shot HOI detection on HICO-DET. We report the performance of unseen classes under three settings (original test set, style-transferred, and degraded images).}
  \label{tab:comparison}
  \resizebox{0.99\textwidth}{!}{
    \begin{tabular}{l c c c c c c c c c c c c}
      \toprule
      \multirow{2}*{Method} 
      & \multicolumn{4}{c}{Original} 
      & \multicolumn{4}{c}{Style-transferred} 
      & \multicolumn{4}{c}{Degraded} \\
      \cmidrule(lr){2-5} \cmidrule(lr){6-9} \cmidrule(lr){10-13}
      & UV & UO & RF-UC & NF-UC & UV & UO & RF-UC & NF-UC & UV & UO & RF-UC & NF-UC \\
      \midrule
      \textbf{\textit{Supervised Methods:}}  \\
      HOICLIP~\cite{ning2023hoiclip} & 24.30 & 16.20 & 25.53 & 26.39 & -- & -- & -- & -- & -- & -- & -- & -- \\
      UniHOI~\cite{cao2023UniHOI} & 22.18 & 13.67 & 23.41 & 26.89 & -- & -- & -- & -- & -- & -- & -- & -- \\
      LogicHOI~\cite{li2024logichoi} & 24.57 & 15.64 & 25.97 & 26.84 & -- & -- & -- & -- & -- & -- & -- & -- \\
      CLIP4HOI~\cite{mao2023clip4hoi} & 26.02 & 31.79 & 28.47 & 31.44 & -- & -- & -- & -- & -- & -- & -- & -- \\
      SICHOI~\cite{luo2024SIC}   & - & - & 34.24 & 34.52 \\
      CMMP~\cite{ting2024CMMP}    & 26.23 & 33.76 &  29.45 & 32.09 & 17.32 & 24.47 & 20.27 & 22.97 & 20.84 & 27.56 & 24.27 & 26.49 \\
      EZ-HOI~\cite{lei2024ezhoi}  & 25.10 & 33.28 & 29.02 & 33.66 & 16.58 & 23.78 & 19.58 & 23.99 & 19.84 & 26.98 & 23.50 & 27.13 \\
      HOLa~\cite{lei2025hola}    & 27.91 & \textbf{36.45} & 30.61 & \textbf{35.25} & 18.83 & 25.72 & 20.66 & \textbf{25.05} & 21.90 & 28.33 & 24.77 & \textbf{27.35} \\
      \textbf{\textit{Weakly-Supervised Methods:}}  \\
      OpenCat~\cite{Zheng_2023_opencat} & 19.48 & 23.84 & 21.46 & 23.25 & -- & -- & -- & -- & -- & -- & -- & --  \\
      \textbf{\textit{Training-Free Method:}}  \\
      AgentHOI & \textbf{29.85} & 33.70 & \textbf{38.64} & 28.04 & \textbf{26.45} & \textbf{28.34} & \textbf{32.55} & 23.05 & \textbf{27.61} & \textbf{28.99} & \textbf{34.79} & 24.56 \\
      \bottomrule
    \end{tabular}
}
\vspace{-1em}
\end{table}

\noindent \textbf{HOI Detection.}
Tab.~\ref{tab:hico_det_dataset} further compares AgentHOI with representative weakly-supervised HOI detection methods on the default HICO-DET setting. Despite requiring no HOI-specific training, AgentHOI achieves the best overall performance (29.61 mAP), outperforming all weakly-supervised counterparts.
More importantly, AgentHOI shows a remarkable advantage on the Rare split, achieving 40.33 mAP, which significantly surpasses previous methods by a large margin. Since rare categories contain very limited training samples, weakly-supervised approaches still struggle to learn reliable visual interaction patterns from scarce data. In contrast, our MLLM-based training-free framework leverages rich vision-language priors acquired from large-scale pretraining, enabling stronger recognition of infrequent or long-tail HOI categories without relying on dataset-specific optimization. This advantage also reflects the effectiveness of our reasoning-and-grounding design: multi-round reasoning helps infer plausible long-tail interactions from scene context, while descriptive grounding improves localization of the corresponding human-object pairs in ambiguous scenes.

\begin{table}[ht]
    \centering
    \begin{minipage}{0.5\textwidth}
        \centering
        \caption{Performance comparison with others of HOI detection on HICO-DET.}
        \label{tab:hico_det_dataset}
        \resizebox{0.99\textwidth}{!}{
        \begin{tabular}{l ccc}
            \toprule
            Method & Full & Rare & Non-Rare \\
            \midrule
            \textbf{\textit{Weakly-Supervised Methods:}}  \\
            Align-Former~\cite{kilickaya2021AlignFormer} & 20.85 & 18.23 & 21.64 \\
            Weakly-HOI~\cite{wan2023weaklyHOI} & 25.70 & 24.52 & 26.05 \\
            OpenCat~\cite{Zheng_2023_opencat}  & 25.82 & 24.35 & 26.19 \\
            \textbf{\textit{Training-Free Method:}}  \\
            AgentHOI & \textbf{29.61} & \textbf{40.33} & \textbf{26.41} \\
            \bottomrule
        \end{tabular}
        }
    \end{minipage}
    \hfill
    \begin{minipage}{0.48\textwidth}
        \centering
        \caption{Effectiveness of the proposed modules. \textit{CMR}: Context-aware Multi-round Reasoning. \textit{MIL}: Multifaceted Interaction Localization.}
        \label{tab:module_comparison}
        \resizebox{0.99\textwidth}{!}{
        \begin{tabular}{lccc}
            \toprule
            Method & Full & Rare & Non-Rare \\
            \midrule
            AgentHOI-Base & 19.44 & 22.00 & 18.68 \\
            + \textit{CMR} & 23.20 & 31.99 & 20.57 \\
            + \textit{MIL} & \textbf{29.61} & \textbf{40.33} & \textbf{26.41} \\
            \bottomrule
        \end{tabular}
        }
    \end{minipage}
\end{table}



\noindent \textbf{Open-vocabulary HOI Detection.}
To further evaluate the generalization ability of AgentHOI beyond HICO-DET, we conduct additional experiments on the challenging SWIG-HOI benchmark, which contains a large and diverse HOI vocabulary (Tab.~\ref{tab:swig_dataset}). On the unseen SWIG-HOI categories, AgentHOI achieves 11.61 mAP without using any HOI-specific training data. This result is competitive with fully supervised methods trained on SWIG-HOI, and even surpasses CMD-SE, which obtains 10.70 mAP. These results suggest that AgentHOI can transfer to unseen HOI categories in another dataset without dataset-specific retraining.

\begin{table}[htbp]
    \centering
    \begin{minipage}[t]{0.48\textwidth}
        \centering
        \caption{Performance comparison with state-of-the-art models on the SWIG-HOI dataset.}
        \label{tab:swig_dataset}
        \resizebox{\linewidth}{!}{
        \begin{tabular}{lcccc}
            \hline
            Method & Full & Rare  & Non-Rare & Unseen \\
            \hline
            \rowcolor{gray!20}  \multicolumn{5}{l}{Fully-supervised Methods} \\
            QPIC~\cite{tamura2021qpic} &  11.12 & 10.84 & 16.95 & 6.21 \\
            GEN-VLKT~\cite{liao2022gen} &  10.87 & 10.41 & 20.91 & - \\
            MP-HOI~\cite{yang2024MP_HOI} & 12.61 & 14.78 & 20.28 & - \\
            THID~\cite{wang2022_THID} &  13.26 & 12.82 & 17.67 & 10.04 \\
            CMD-SE~\cite{lei2024CMD-SE} & 15.26 & 14.64 & 21.46 & 10.70 \\
            SGC-Net~\cite{lin2025sgc} & \textbf{17.20} & \textbf{16.55} & \textbf{23.67} & \textbf{12.46} \\
            \rowcolor{gray!20}  \multicolumn{5}{l}{Training-Free Methods} \\
            AgentHOI-Base &  9.64 & 9.35 & 10.24 & 8.74  \\
            AgentHOI      & \textbf{13.00} & \textbf{13.01} & \textbf{14.45} & \textbf{11.61} \\
            \hline
        \end{tabular}
        }
    \end{minipage}
    \hfill
    \begin{minipage}[t]{0.48\textwidth}
        \centering
        \caption{Effectiveness of Context-aware Multi-round Reasoning. IHI: Initial HOI Identification, HR: HOI Re-mining, AR: Action Reassignment.}
        \label{tab:performance_comparison}
        \resizebox{\linewidth}{!}{
        \begin{tabular}{llccc}
            \toprule
            Method & Setting & Precision & Recall & F1 \\
            \midrule
            IHI       & object & 74.82 & 85.69 & 78.01 \\
            IHI+HR    & object & 79.05 & 92.31 & 83.15 \\
            IHI+HR+AR & object & \textbf{79.05} & \textbf{92.31} & \textbf{83.15} \\
            \midrule
            IHI       & action & 59.66 & 43.94 & 46.88 \\
            IHI+HR    & action & \textbf{64.63} & 48.48 & 51.34 \\
            IHI+HR+AR & action & 63.07 & \textbf{76.16} & \textbf{65.72} \\
            \midrule
            IHI        & HOI  & 54.36 & 39.84 & 42.29 \\
            IHI+HR     & HOI  & \textbf{59.77} & 44.89 & 47.21 \\
            IHI+HR+AR  & HOI  & 58.76 & \textbf{72.51} & \textbf{64.91} \\
            \bottomrule
        \end{tabular}
        }
    \end{minipage}
\end{table}

\subsection{Ablation Study}

\noindent \textbf{Effectiveness of the Proposed Modules.}
We conduct an ablation study to evaluate the effectiveness of the Context-aware Multi-round Reasoning~(CMR) and Multifaceted Interaction Localization~(MIL). As shown in Tab.~\ref{tab:module_comparison}, adding CMR significantly enhances interaction recognition, yielding an overall mAP improvement of 3.76. Building upon this, the integration of the MIL further refines the model's ability to differentiate instances involved in similar interactions. Notably, the combined approach achieves a remarkable mAP of 29.61 on the Full categories. These results highlight the effectiveness of AgentHOI for in-the-wild training-free HOI detection.

\noindent\textbf{Effectiveness of Context-Aware Multi-Round Reasoning.} 
We evaluate the reasoning process using image-level recognition metrics (Tab.~\ref{tab:performance_comparison}):
(1) \textit{Object recognition (Line 1--3):} Introducing HR improves Precision/Recall/F1 by 4.23\%/6.62\%/5.14\%, showing it effectively leverages anchor HOI instances as context to detect more valid HOIs. 
(2) \textit{Action recognition (Line 4--6):} Adding HR alone outperforms IHI by detecting more missing HOIs. Adding AR further improves the F1 score by a large margin (14.38\%), highlighting the benefit of fine-grained action priors for context-aware prediction.
(3) \textit{HOI recognition:} Both HR and AR contribute complementary gains—HR for better object context, AR for refined actions—jointly validating the effectiveness of our multi-round reasoning framework.


\noindent \textbf{Comparison of Strategies for Interaction Localization.} 
Tab.~\ref{tab:strategy_comparison} compares three strategies: ``Description-only'' (semantic descriptions), ``Relation-only'' (verb-centric phrases), and their combination.
The ``Description-only'' strategy outperforms ``Relation-only'' by 3.40  in overall mAP, suggesting that attribute-rich descriptions provide stronger grounding than verbs alone. We observe that ``Relation-only'' performs poorly in both SingleHOI and MultiHOI cases, likely due to verb ambiguity without object context.
Combining both strategies yields the best results, confirming their complementary strengths in interaction localization.


\begin{table}[ht]
    \centering
    \begin{minipage}{0.56\textwidth}
        \centering
        \caption{Comparison of strategies for multifaceted interaction localization. S-HOI: SingleHOI, M-HOI: MultiHOI.}
        \label{tab:strategy_comparison}
        \resizebox{0.99\textwidth}{!}{
        \begin{tabular}{lccccc}
        \toprule
        Strategy & Full & Rare & Non-Rare & S-HOI & M-HOI \\
        \midrule
        Description-only & 27.70 & 35.85 & 25.26 & 42.46 & 24.18\\
        Relation-only     & 24.30 & 31.09 & 22.27 & 38.48 & 21.04 \\
        Multifaceted      & \textbf{29.61} & \textbf{40.33} & \textbf{26.41} & \textbf{44.28} & \textbf{26.41} \\
        \bottomrule
        \end{tabular}
        }
    \end{minipage}
    \hfill
    \begin{minipage}{0.42\textwidth}
        \centering
        \caption{Effect of multiple HOI re-mining iterations.}
        \label{tab:remining_iterations}
        \resizebox{0.99\textwidth}{!}{
        \begin{tabular}{lccc}
        \toprule
        \# (Iterations) & Precision & Recall & F1-score \\
        \midrule
        One   & \textbf{58.76} & 72.51 & \textbf{64.91} \\
        Two   & 58.02 & 72.97 &  64.64 \\
        Three & 57.21 & \textbf{73.67} & 64.40 \\
        \bottomrule
        \end{tabular}
        }
    \end{minipage}
\end{table}

\noindent \textbf{Effect of Multiple HOI Re-mining Iterations}
As shown in Tab.~\ref{tab:remining_iterations}, we evaluate the impact of applying one, two, and three re-mining iterations. The results indicate that a single re-mining iteration is sufficient for discovering meaningful interactions. While additional iterations improve recall, they lead to a slight drop in precision and overall F1-score, suggesting diminishing returns and potential over-generation of interaction candidates.

\noindent \textbf{Compatibility with Open-source MLLM.} 
To evaluate the adaptability of AgentHOI, we further tested it with powerful open-source large vision-language model Qwen2.5-VL-72B. As shown in Tab. ~\ref{tab:lvlm_compatibility}, while GPT-4o achieves the best performance (29.61 mAP), Qwen2.5-VL-72B also delivers strong results (26.12 mAP), surpassing SOTA weakly supervised HOI detectors (25.82 mAP). This shows AgentHOI’s adaptability across both proprietary and open-source MLLMs.

\noindent \textbf{Effect of the balancing coefficient $\lambda$.}
We further analyze the influence of the balancing coefficient $\lambda$, which controls the relative contribution of the grounding reliability when computing the final HOI confidence score. As shown in Tab.~\ref{tab:lambda}, moderate values of $\lambda$ consistently yield better performance. When $\lambda$ is too small (e.g., $\lambda=0.1$), the final confidence relies primarily on semantic classification scores, making the model less sensitive to localization reliability. Increasing $\lambda$ to $0.3$ improves both Full and Rare mAP, indicating that incorporating grounding confidence helps suppress spatially inconsistent predictions. However, further increasing $\lambda$ (e.g., $\lambda=0.5$) slightly degrades performance, suggesting that overly emphasizing localization scores may introduce noise from imperfect grounding results.

\begin{table}[ht]
    \centering
    \begin{minipage}{0.46\textwidth}
        \centering
        \caption{Compatibility with different MLLMs}
        \label{tab:lvlm_compatibility}
        \begin{tabular}{lccc}
            \hline
            MLLM & {Full} & {Rare} & {Non-Rare} \\
            \hline
            Qwen2.5-VL-72B~\cite{bai2025qwen2-5} & 26.12 & 35.00 & 23.47 \\
            GPT-4o~\cite{achiam2023gpt} & \textbf{29.61} & \textbf{40.33} & \textbf{26.41} \\
            \hline
        \end{tabular}
    \end{minipage}
    \hfill
    \begin{minipage}{0.52\textwidth}
        \centering
        \caption{Ablation study on the balancing coefficient $\lambda$ in the final confidence score.}
        \label{tab:lambda}
        \begin{tabular}{lccc}
            \hline
            $\lambda$ & Full & Rare & Non-Rare \\
            \hline
            0.1 & 29.42 & 39.90 & 26.29 \\
            0.3 & \textbf{29.61} & \textbf{40.33} & \textbf{26.41} \\
            0.5 & 29.44 & 39.58 & 26.41 \\
            \hline
        \end{tabular}
    \end{minipage}
\end{table}

\subsection{Qualitative Results}
\label{suppsec:ana_qualitative}


\begin{figure}[t]
    \centering
    \begin{subfigure}{0.51\textwidth}
        \centering
        \includegraphics[width=\textwidth]{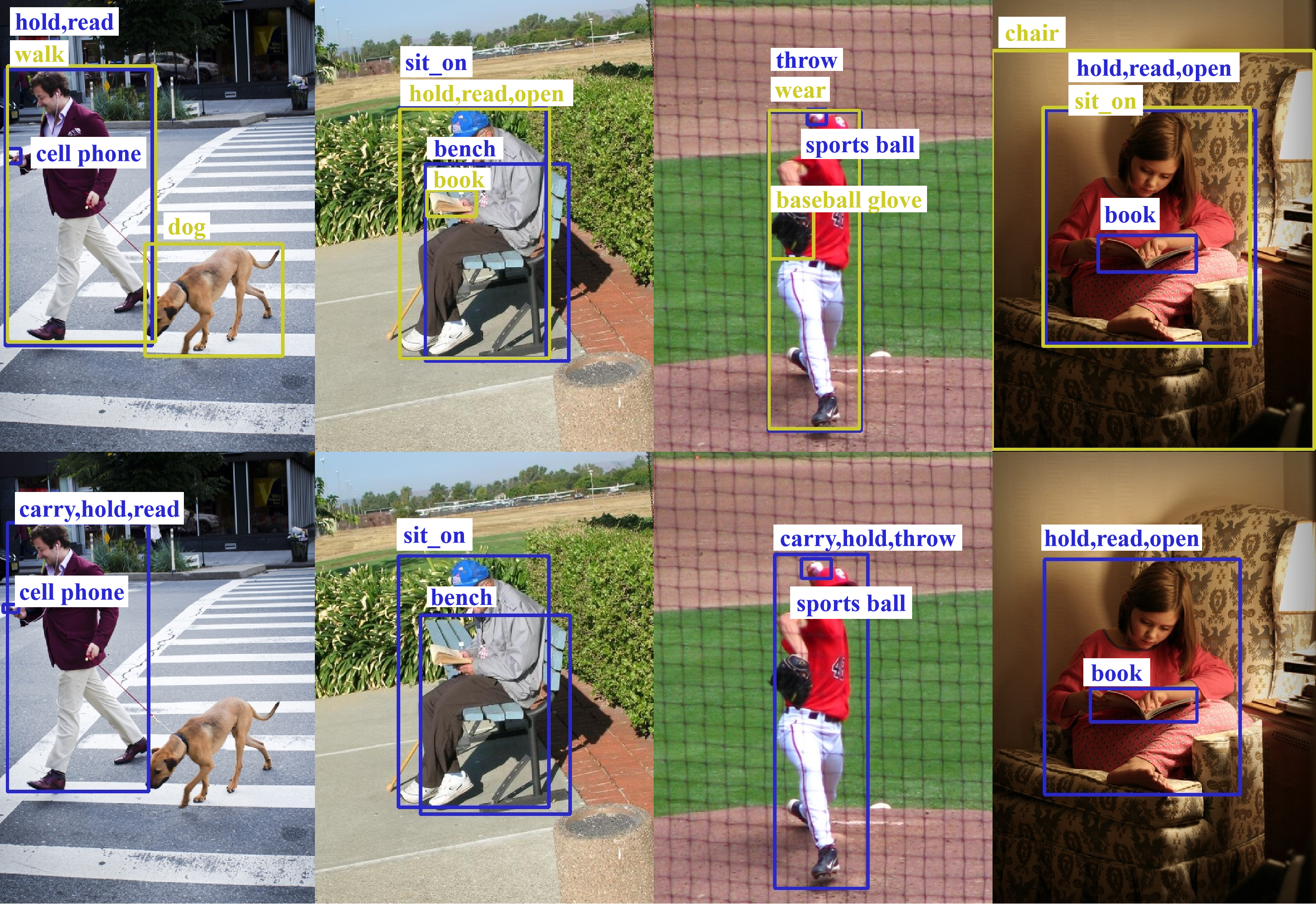}
        \caption{AgentHOI prediction~(first line) vs annotations~(second line).}
        \label{fig:AgentHOIvsGT}
    \end{subfigure}
    \hfill
    \begin{subfigure}{0.48\textwidth}
        \centering
        \includegraphics[width=\textwidth]{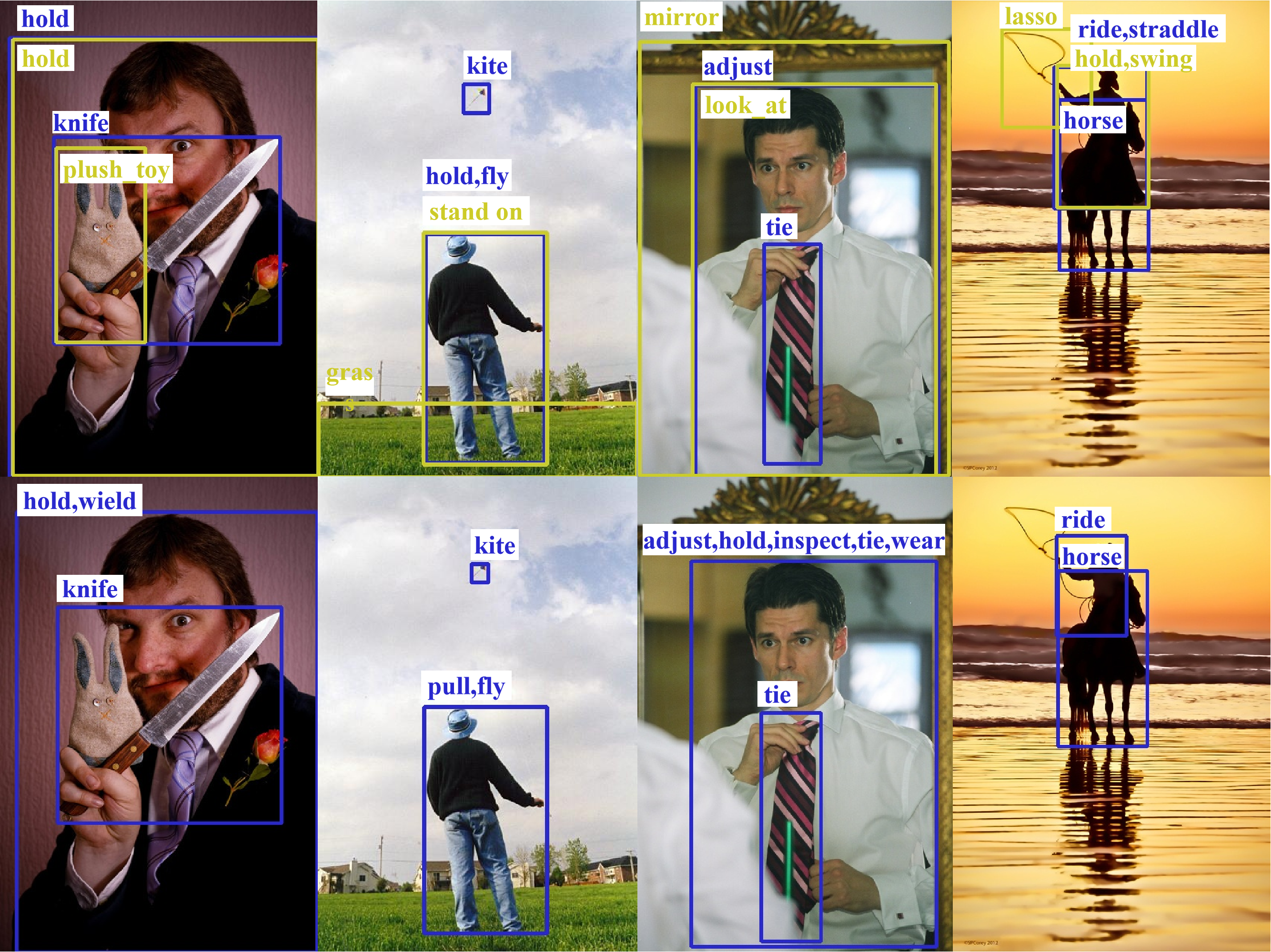}
        \caption{AgentHOI prediction w/o predefined categories~(first line) vs annotations~(second line).}
        \label{fig:NoActAndObjvsGT}
    \end{subfigure}
    \caption{Qualitative results on HICO-DET.}
    \label{fig:combined_comparison}
\end{figure}

We first assess the capabilities of AgentHOI through qualitative comparisons with the ground-truth annotations. 
As shown in Fig.~\ref{fig:AgentHOIvsGT}, some images appear to contain incomplete annotations—a common limitation of large-scale HOI datasets, particularly for subtle or overlapping interactions. 
For example, in the first image on the left, the ``walking dog'' interaction is not annotated, resulting in incomplete ground truth. 
In such cases, AgentHOI is able to identify these unlabeled interactions through multi-round reasoning and contextual understanding, demonstrating its robustness and potential to complement existing datasets by revealing plausible yet unannotated interactions.

Beyond this analysis, we further evaluate AgentHOI's generalization in real-world scenarios by examining its performance without leveraging any dataset-specific prior knowledge, including the object set \( \mathcal{C}_o \) and the action set \( \mathcal{C}_a \). 
Notably, AgentHOI represents the \textit{first generative-model-based} approach in the HOI community capable of predicting arbitrary objects and actions, without relying on a predefined vocabulary tied to a specific dataset.
As illustrated in Fig.~\ref{fig:NoActAndObjvsGT}, we compare AgentHOI's predictions with the ground-truth annotations on the HICO-DET test set. 
The results indicate that AgentHOI effectively detects interactions involving actions or objects that are \textit{not included} in the dataset’s predefined categories. 
For instance, in the last row of Fig.~\ref{fig:NoActAndObjvsGT}, where the ground truth only includes the interaction ``ride horse'', AgentHOI not only predicts multiple semantically related actions such as ``ride'' and ``straddle'', but also uncovers additional interactions like ``hold'' and ``swing lasso''. 
These examples highlight the model’s open-vocabulary reasoning capability and its potential to capture a broader and more diverse spectrum of human-object interactions beyond dataset constraints.

\subsection{Inference Cost Analysis}
AgentHOI introduces higher inference cost compared to conventional supervised HOI detectors, since it performs MLLM-based open-vocabulary reasoning instead of a single forward pass with task-specific classifiers. Note that the MLLM reasoning is conducted at the image level, rather than independently for each human-object pair. Therefore, the number of MLLM calls remains independent of the number of candidate pairs, while pair-level processing is mainly handled by the grounding module. Nevertheless, AgentHOI does not introduce any HOI-specific trainable parameters. It coordinates off-the-shelf foundation models, including GPT-4o, whose model size is undisclosed, and GroundingDINO ($\sim$175M parameters). In comparison, a representative supervised HOI detector such as HOLa has around $\sim$195M parameters and takes about 0.1s per image.

This additional inference cost should be viewed as a trade-off for avoiding HOI-specific training and fine-tuning, which typically require costly human-object pair annotations and retraining when new categories or domains are introduced. Therefore, AgentHOI is not intended to replace highly optimized supervised detectors in latency-sensitive settings. Rather, it provides a complementary option for open-world or data-scarce scenarios where flexibility, robustness under distribution shifts, and generalization to unseen HOI categories are important. In these cases, the increased inference latency represents a practical limitation, but it is partly offset by the reduced dependence on task-specific training data.

\section{Conclusion}
\label{sec:conclusion}

We propose AgentHOI, a training-free agentic approach for HOI detection in the wild, leveraging the strengths of MLLMs. Unlike traditional methods constrained by high annotation costs and poor generalization, our approach utilizes MLLM’s robust visual understanding to identify HOI triplets, while addressing the limitations in localization through grounding models. To mitigate the challenges posed by complex scenes with multiple interactions and the difficulty of distinguishing visually similar instances, we introduce context-aware, multi-round reasoning for iterative instance recognition and multifaceted prompts to enhance localization accuracy. Experiments on HICO-DET and SWIG-HOI demonstrate significant improvements in performance, validating the effectiveness and adaptability of AgentHOI in in-the-wild HOID tasks.

\section*{Acknowledgements}
This work was supported by the grants from the National Natural Science Foundation of China (62372014, 62525201, 62132001, 62432001), Beijing Nova Program and Beijing Natural Science Foundation (4252040, L247006).

\bibliographystyle{splncs04}
\bibliography{main}
\end{document}